%% file: main.tex
\documentclass[10pt,twocolumn,letterpaper]{article}

\usepackage[pagenumbers]{cvpr} 

\usepackage{amsmath}
\usepackage{amssymb}
\usepackage{mathtools}
\usepackage{amsthm}
\usepackage{mdframed}
\usepackage{multicol,multirow}
\usepackage{colortbl}
\usepackage{caption}
\usepackage[most]{tcolorbox}

\newtcolorbox{promptbox}[1][]{
    colback=gray!20!white,
    colbacktitle=white,
    coltitle=black,
    colframe=black!75!black,
    boxrule=0.7pt,
    halign title=center,
    title=\textbf{#1}
}

\input{preamble}

\definecolor{cvprblue}{rgb}{0.21,0.49,0.74}
\usepackage[pagebackref,breaklinks,colorlinks,allcolors=cvprblue]{hyperref}

\def\paperID{****} 
\def\confName{CVPR}
\def\confYear{2026}

\title{SemiNFT: Learning to Transfer Presets from Imitation to Appreciation via Hybrid-Sample Reinforcement Learning}

\author{
Melany Yang$^{1, 2}$, Yuhang Yu$^{1}$, Diwang Weng$^{1}$, Jinwei Chen$^{1}$, Wei Dong$^{1\dagger}$\\
$^1$vivo Mobile Communication Co. Ltd, $^2$Zhejiang University
}

\begin{document}
\twocolumn[{
\renewcommand\twocolumn[1][]{#1}
\maketitle
\begin{center}
    \vspace{-5pt}
    \includegraphics[width=0.97\textwidth]{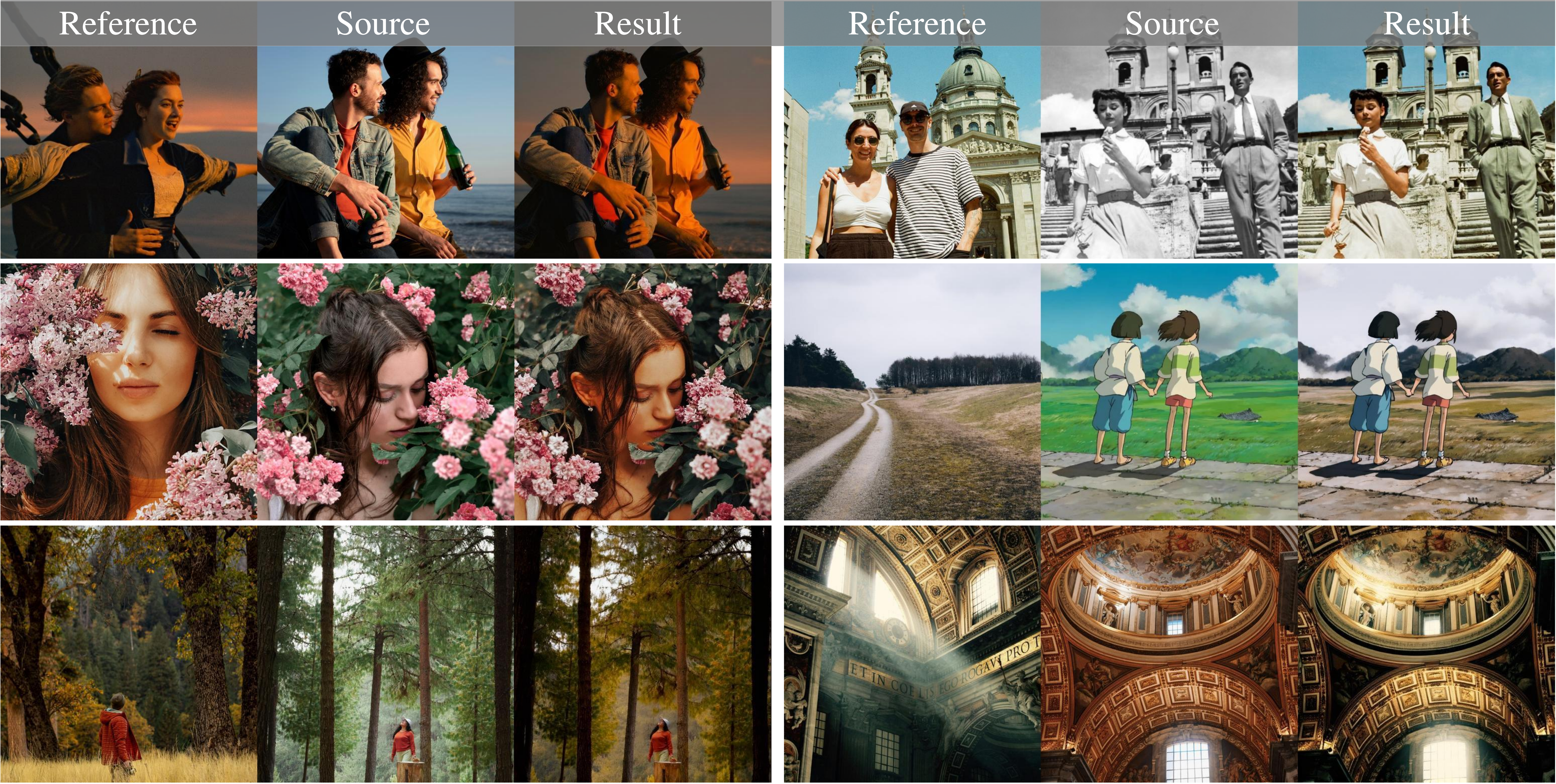}
    \captionsetup{type=figure}
    \vspace{-10pt}
    \caption{%
        \textbf{Photorealistic Preset Transfer Results}. 
        Our preset transfer framework supports diverse practical scenarios, including standard photorealistic preset transfer, anime-reference preset transfer, and even old black-and-white photo colorization, demonstrating its versatility and effectiveness in various color-related tasks.
    }
    \label{homepage}
    \vspace{8pt}
\end{center}
}]

\footnotetext[2]{Corresponding author.}

\begin{abstract}
Photorealistic color retouching plays a vital role in visual content creation, yet manual retouching remains inaccessible to non-experts due to its reliance on specialized expertise. Reference-based methods offer a promising alternative by transferring the preset color of a reference image to a source image. However, these approaches often operate as novice learners, performing global color mappings derived from pixel-level statistics, without a true understanding of semantic context or human aesthetics. 
To address this issue, we propose SemiNFT, a Diffusion Transformer (DiT)-based retouching framework that mirrors the trajectory of human artistic training: beginning with rigid imitation and evolving into intuitive creation.
Specifically, SemiNFT is first taught with paired triplets to acquire basic structural preservation and color mapping skills, and then advanced to reinforcement learning (RL) on unpaired data to cultivate nuanced aesthetic perception. Crucially, during the RL stage, to prevent catastrophic forgetting of old skills, we design a hybrid online-offline reward mechanism that anchors aesthetic exploration with structural review.
Extensive experiments show that SemiNFT not only outperforms state-of-the-art methods on standard preset transfer benchmarks but also demonstrates remarkable intelligence in zero-shot tasks, such as black-and-white photo colorization and cross-domain (anime-to-photo) preset transfer. These results confirm that SemiNFT transcends simple statistical matching and achieves a sophisticated level of aesthetic comprehension. Our project can be found at \url{https://melanyyang.github.io/SemiNFT/}.
\end{abstract}

\section{Introduction}
\label{Introcution}
Photorealistic color retouching plays an essential role in the post-production of photography, where color tone and visual consistency critically influence the final aesthetic and perceptual quality of images.
Early approaches relied on global statistical alignment, such as histogram matching \cite{reinhard2001color, pitie2005, rabin2010}, while subsequent learning-based methods leveraged deep neural networks to learn color mappings for precise color retouching \cite{Gong2025salut, ke2023neural, yoo2019photorealistic}. 

However, we identify three core challenges in existing methods that hinder precise and controllable photorealistic preset transfer:
\textbf{(I) Semantic-Agnostic Mapping:} Existing methods predominantly learn global color mappings for the source image. These methods fail to account for local semantic contexts, i.e., the distinct functional regions and object categories within a scene, consequently leading to appearance adjustments that lack spatial precision and semantic coherence.
\textbf{(II) Data Scarcity:} Data-driven approaches require well-designed retouching pairs, yet such datasets for photorealistic preset transfer remain scarce.
\textbf{(III) Misalignment with Perception:} Current benchmarks predominantly rely on minimizing the discrepancy between predicted outputs and human-retouched targets, which inherently limits model performance to the subjective quality of the training data.

In this paper, we systematically address the above challenges by designing a curriculum-style training paradigm inspired by the learning process of a human retouching expert.
Specifically, we propose \textbf{SemiNFT}, a semi-offline, samples-enabled, negative-aware fine-tuning framework for photorealistic color retouching. Built upon a diffusion transformer (DiT) backbone, SemiNFT takes a source image and a reference image as multi-input conditions, enabling photorealistic transfer of the reference preset onto the source image, as illustrated in Fig.~\ref{homepage}.
The core of SemiNFT lies in the curriculum-style learning strategy inspired by the way human retouching experts acquire aesthetic competence. That is, SemiNFT is designed to progress from rigid imitation to intuitive creation. Following this principle, the process starts with cold-start supervised fine-tuning on paired image triplets to capture fundamental structural relationships, and subsequently transitions to reinforcement learning (RL) on unpaired data to cultivate higher-level aesthetic perception.

Different from most RL paradigms that cannot effectively leverage negative data, we perform negative-aware finetuning \cite{zheng2025diffusionnft} directly on the forward process via flow matching.
To avoid potential reward hacking \cite{NIPS2017_32fdab65}, and prevent the model from forgetting the structural preservation skills learned in the cold-start stage or overfitting to purely aesthetic preferences, we propose a hybrid online-offline reward mechanism to review old skills. 
Beyond online rollouts, we also include a small set of offline samples serving as anchor points to regularize the reward model aligned with human perceptual judgments.

To facilitate training, we carefully construct a specified dataset comprising $3,200$ paired image triplets sourced from curated video clips and Qwen-Image-Edit \cite{qwenimageedit2025}-guided generation, $1,500$ unpaired images (source, reference) collected through image-to-image retrieval. 
While for performance evaluation, we introduce a standardized evaluation protocol based on multiple Vision-Language Models (VLMs), including Gemini-2.5-flash \cite{gemini2.5}, GPT-4o \cite{gpt}, and Qwen3-VL-32B \cite{qwen3vl},  facilitating comprehensive and objective assessment. 

With the proposed curriculum-style training pipeline, dataset construction, and hybrid reward mechanism, SemiNFT demonstrates strong and consistent performance across a wide range of preset transfer scenarios. As illustrated in Fig.~\ref{homepage}, beyond standard settings where both the source and reference images are realistic color photographs, SemiNFT generalizes effectively to more challenging cases, including black-and-white photo colorization and cross-domain preset transfer.

We summarize our key contributions as follows:
\begin{itemize}
\item We propose a DiT-based framework for photorealistic preset transfer that features an efficient two-stage training pipeline: a supervised cold-start phase on paired triplets followed by an RL phase on unpaired data.
\item We design a hybrid online-offline reward mechanism to prevent catastrophic structural errors and misalignment with human perception.
\item We build and release a diverse dataset and benchmark for photographic preset transfer, along with a standardized VLM-based evaluation protocol that facilitates multi-dimensional performance assessment.
\end{itemize}

\section{Related Work}
\subsection{Photorealistic Preset Transfer}
Photorealistic preset transfer aims to modify the color distribution of a target image to match that of a reference image while preserving the underlying content structure. Early approaches focused on global color statistics, such as histogram matching and moment alignment in perceptually uniform color spaces \cite{reinhard2001color, pitie2005, rabin2010}. 
While effective for simple scenes, these methods operate primarily at a global level and lack semantic awareness, often producing artifacts when the reference and target images differ significantly in content or structure.
With the advent of deep learning, subsequent methods introduced data-driven representations to model more complex color transformations beyond global statistics \cite{luan2017deep, li2017universal, yoo2019photorealistic, ke2023neural, Gong2025salut}. 
For example, \cite{yoo2019photorealistic} introduced wavelets into the design of the network to reduce the noise amplification of the image. \cite{ke2023neural} designed a Deterministic Neural Color Mapping module to consistently operate on each pixel. \cite{Gong2025salut} proposed a spatially adaptive 4D Look-Up Table (LUT) model to enable context-sensitive color adjustments.
Despite these advances, learning-based preset methods typically rely on global or weakly localized statistics and struggle to simultaneously maintain color fidelity, structural preservation, and semantic consistency. Moreover, most approaches lack the capacity to flexibly leverage reference images in open-domain and generative settings.

\subsection{Universal Image Editing}
Diffusion models have recently emerged as a powerful paradigm for high-fidelity image generation \cite{ho2020ddpm, song2021scorebased, rombach2022high}. By modeling a gradual denoising process from noise to data, diffusion-based methods achieve superior visual quality and training stability compared to adversarial approaches. 
Recent advances have further explored Diffusion Transformer-based architectures (DiT) in text-to-image generation, including SD series \cite{sd3.5}, Flux.1 series \cite{flux2024, fluxkontex}, Qwen-Image series \cite{qwenimage2025, qwenimageedit2025}, which improve scalability and conditioning expressiveness for image generation and editing. 
Beyond text-based image generation, recent unified image editing models, which edit the input image with a given text prompt, have demonstrated strong capabilities in both visual understanding and image editing. 
Closed-source systems such as GPT-4o \cite{gpt} and Nano Banana \cite{gemini2.5} exhibit the ability to perform generalized image editing tasks, including color adjustment, under diverse forms of multimodal instruction. 
While these models highlight the potential of unified comprehension–generation architectures for image retouching, their retouching behavior is often implicit and entangled with other editing objectives.

\section{Method}
\begin{figure*}[h]
  \centering
  \includegraphics[width=\linewidth]{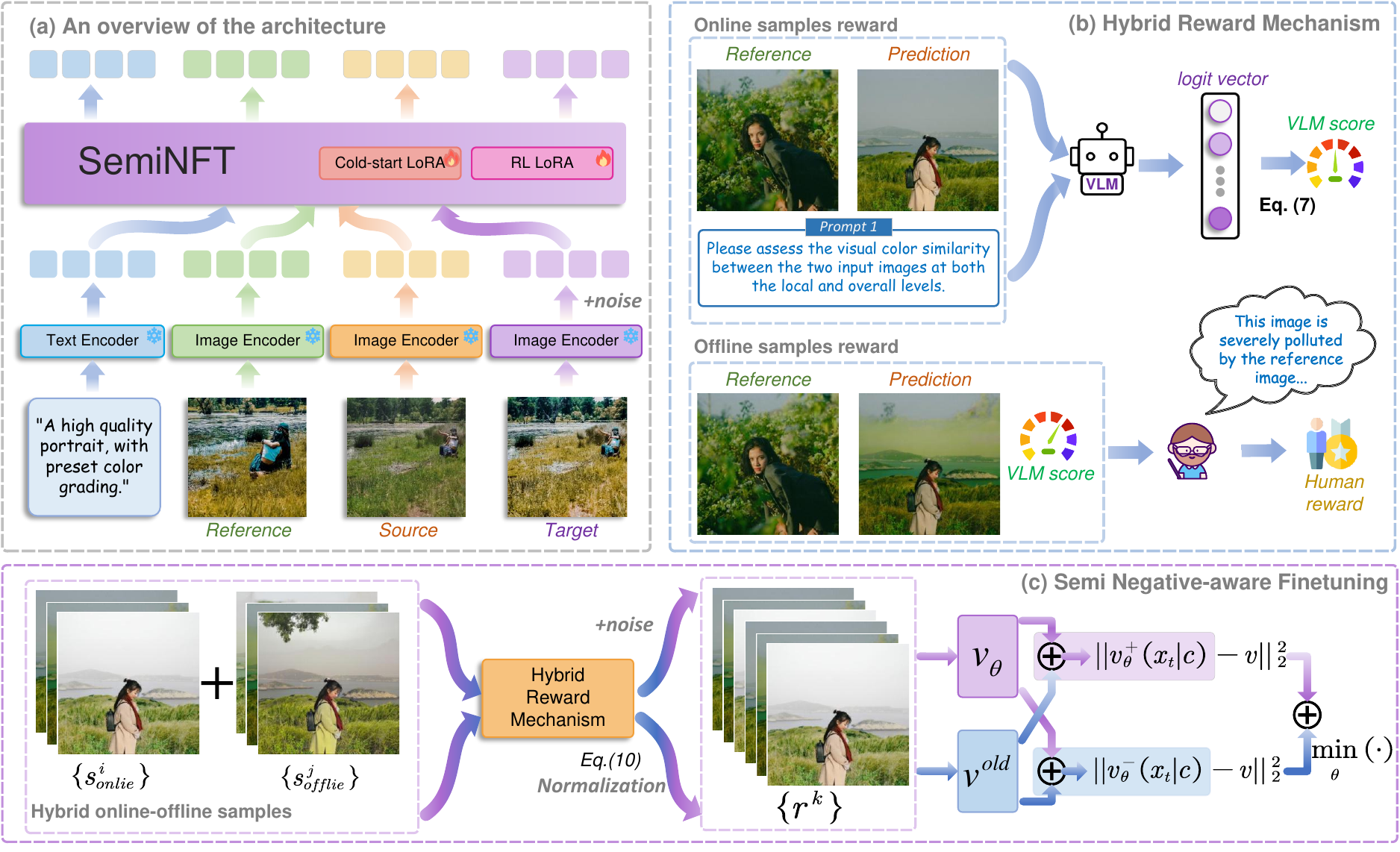}
  \caption{Illustration of the SemiNFT. (a) The architecture integrates reference and source images as inputs through causal attention. The cold-start and post-training LoRA modules are trained in the cold-start and post-training stage, respectively. (b) The scoring process of SemiNFT includes online reward and offline reward. (c) We fine-tune the velocity predictor using DiffusionNFT, enhanced by incorporating a prepared SFT dataset.}
  \label{overview}
  \vspace{-0.5em}
\end{figure*}

\subsection{Preliminary}


Rectified Flow ~\cite{liu2022flow} defines intermediate samples $x_t$ as $x_t = (1-t)x_0 + tx_1, t \in [0,1]$,
where $x_0$ is sampled from the true data distribution and $x_1$ is a noise sample.
As for inference, the iterative denoising process in flow matching models can be formulated as a Markov Decision Process \cite{black2024training}. The state at step t is defined as $s_t = (c, t, x_t)$, and the action $a_t$ corresponds to the denoised sample $s_0 = (c, T, x_T)$, where $x_T \sim \mathcal{N(\mathbf{0}, \mathbf{I})}$. The reward is only provided at the final step: $R(s_t, a_t) = r(x_0, c)$ if $t = 0$, and $0$ otherwise.

With this formulation, many recent RL algorithms~\citep{shao2024deepseekmath, liu2025flow, xue2025dancegrpo} are built upon the policy gradient framework, such as PPO and GRPO. For example, DiffusionNFT~\citep{zheng2025diffusionnft} optimizes diffusion models directly on the forward process via flow matching. 
It contrasts positive and negative generations to define an implicit policy improvement direction, and defines a contrastive improvement direction between the two policies learned on them:
\begin{equation}\label{eq:policy_opt}
\begin{aligned}
\mathcal{L}(\theta) &= \mathbb{E}_{c,\pi^{\mathrm{old}}(x_0\mid c),t}
\Big[ r\, \|v^+_\theta(x_t,c,t)-v\|_2^2 \\
&+ (1-r)\, \|v^-_\theta(x_t,c,t)-v\|_2^2 \Big],
\end{aligned}
\end{equation}
where $v$ is the target velocity field, $v^+_\theta$ and $v^-_\theta$ denote respectively the implicit positive and negative policies that are combinations of the old policy $v^{old}$ and the training policy $v_\theta$, weighted by a hyperparameter $\beta$:
\begin{align}
v^+_\theta(x_t,c,t)&:=(1-\beta)\,v^{old}(x_t,c,t)+\beta\,v_\theta(x_t,c,t), \\
v^-_\theta(x_t,c,t)&:=(1+\beta)\,v^{old}(x_t,c,t)-\beta\,v_\theta(x_t,c,t),
\end{align}
where $r \in [0, 1]$ is the reward.


\subsection{Overview of SemiNFT}
As illustrated in Fig.~\ref{overview}, it takes a source image and a reference image as multi-input conditions. The framework consists of three tightly coupled components:
(i) a curriculum-style two-stage training pipeline that mirrors the progression of human retouching expertise;
(ii) a hybrid online–offline reward mechanism designed to mitigate reward hacking and preserve structural fidelity; and
(iii) a purpose-built dataset construction pipeline tailored to the requirements of each training stage.

We adopt a progressive two-stage training strategy to gradually improve model controllability and alignment with human preferences. In the first stage, also referred to as the cold-start stage, we introduce a cold-start LoRA module and perform supervised fine-tuning on a paired dataset, where ground-truth target images are available. This stage enables the base DiT model to learn the structural correspondence among the source, reference, and target images, thereby establishing a solid foundation for subsequent reinforcement learning. In the second stage, we freeze the cold-start LoRA and train an additional post-training LoRA module using the negative-aware fine-tuning algorithm on a high-quality unpaired dataset. This module is to further refine output quality in the absence of direct supervision.

Beyond the training paradigm, our framework incorporates two critical designs to improve generalization and training reliability:
(1) A staged dataset construction pipeline, where paired data support structural learning in Stage 1 and unpaired data enable preference-driven refinement in Stage 2;
(2) A hybrid reward design that combines offline human-rated scores with online feedback. Specifically, we curate a small validation set of diverse samples and assign them human-derived quality scores. These scores serve as anchor points to regularize the reward model, discouraging degenerate behaviors such as excessive sample generation.

\subsection{Stage 1: Cold-Start Training}
We build our model upon the FLUX.1-[dev] \cite{flux2024} and introduce a dedicated LoRA module, which we refer to as the \textit{cold-start LoRA}. In this stage, we train the cold-start LoRA for $10,000$ steps exclusively on the cold-start dataset, a collection of quadruplets comprising a textual description, a source image, a reference image, and the corresponding target image (see Section \ref{cold-start dataset} for details). To prevent cross-contamination between input modalities during training, we modify the original bidirectional attention in the DiT blocks to causal attention with a structured mask to ensue the generation remains conditioned on both images without leaking information across branches. 
Specifically, tokens from the reference and source image branches are only allowed to attend within their own branches and are explicitly blocked from attending to the main branch, which contains noisy latent tokens and text conditioning tokens. Conversely, the main branch follows standard causal attention and can attend to all tokens, ensuring that generation remains conditioned on both images without leaking information across branches.

\subsection{Stage 2: Reinforcement Learning with Hybrid Samples}
Despite the ability to establish basic alignment, SFT on the cold-start dataset faces significant limitations in professional preset transfer scenarios. High-fidelity paired data, in which the target image is meticulously color graded to match the reference, are extremely scarce because they require expert-level manual adjustments and are prohibitively expensive to produce. Consequently, relying solely on such limited ``perfect" examples constrains the model’s ability to generalize to diverse real-world conditions, including variations in lighting, contrast, and color grading styles.
To overcome these constraints and push the boundaries of both color fidelity and aesthetic quality, we transition to a reinforcement learning paradigm in the second stage. Building upon the framework of DiffusionNFT \citep{zheng2025diffusionnft}, we integrate a VLM-driven reward model with a small set of offline samples. This hybrid approach enables the model to optimize for both preset alignment and photorealism, significantly mitigating reward hacking.

\subsubsection{VLM-Based Reward Modeling} \label{VLMreward}
We employ a pretrained VLM, Qwen3-VL-8B-Instruct \cite{qwenimage2025}, as a reward model to assess the quality of preset transfer. The reward is designed to assess preset similarity at both global and local levels.
Given a reference image $I_{\text{ref}}$ and a predicted image $I_{\text{pred}}$, we form the input pair 
$\mathbf{X}=(I_{\text{ref}}, I_{\text{pred}})$. To obtain a deterministic and interpretable score, we prompt the VLM with a carefully designed textual instruction (see Appendix \ref{template}) that constrains the output to a single categorical token from a predefined scoring set $\mathcal{M}$, e.g., tokens representing scores from $1$ to $5$. Let $w(r)$ denote the numerical value associated with token $r \in \mathcal{M}$. The expected similarity score is computed as:
\begin{align}
    s_{\text{raw}}(\mathbf{X}) = \sum_{r \in \mathcal{M}} w(r) \cdot p(R = r \mid \mathbf{X}),
    \label{eq:reward_sim}
\end{align}
where $p(R = r \mid \mathbf{X})$ is derived from the softmax over the final logit vector of the VLM. This formulation captures the model’s confidence distribution over possible ratings. 



\subsubsection{Hybrid online-offline samples.}
While RL from human feedback enables the model to align with perceptual preferences, training solely on online samples can lead the model to drift away from realistic image structures and excessive preset transfer. To address this, we incorporate a small set of human-labeled offline samples during the second-stage training.
Specifically, at each training step, we construct a hybrid batch consisting of two types of samples:
(1) \textit{Online samples}: generated on-the-fly by the current policy, i.e., the DiT backbone with frozen cold-start LoRA and trainable RL LoRA;
(2) \textit{Offline samples}: drawn from a small validation set of diverse samples.
For online samples, we compute rewards using the VLM-based reward model described in Section \ref{VLMreward}. For offline samples, we assign each sample a fixed score via human labeling, which serves as anchor points to regularize the reward model, discouraging degenerate behaviors such as excessive sample generation. For example, a VLM-approved may not align with the human preference, thus forced to a lower reward. During training, the policy is aware of reducing the probability of the occurrence of artificially suppressed samples, thereby avoiding excessive transferring.

This hybrid approach provides two key benefits. First, the offline samples act as anchor points in the policy space, preventing catastrophic forgetting of structural priors learned during the cold-start stage. Second, they regularize the reward landscape, reducing the risk of the policy exploiting artifacts or shortcuts in the VLM reward model. 

\subsubsection{SemiNFT Pipeline}
As illustrated in Fig. \ref{overview}, our training pipeline consists of three iterative steps: sampling, scoring, and policy fine-tuning, which collectively steer the model toward generating well-transferred and preference-aligned outputs.
At each iteration, we first perform a rapid rollout for a given source-reference pair, generating an online sample set $\mathcal{X}_{\text{online}}$ of $G_{\text{online}}$ samples sampled from the policy $\pi_{\text{old}}$, with the raw reward scored by the VLM. Simultaneously, we include an offline set $\mathcal{X}_{\text{offline}}$ of $G_{\text{offline}}$ offline samples accompanied by human-labeled raw scores. The combined set $\mathcal{X}=\mathcal{X}_{\text{online}} \cup \mathcal{X}_{\text{offline}}$ of $G = G_{\text{online}} + G_{\text{offline}}$ samples is then used to compute normalized optimality rewards via group-wise standardization:
\begin{equation}
r(x_0^i) = \frac{1}{2} \cdot \mathrm{clip}\left( \frac{s_{\text{raw}}(x_0^i) - \mu}{\sigma},\ -1,\ 1 \right) + \frac{1}{2},
\label{eq:reward_norm}
\end{equation}
where $\mu$ and $\sigma$ denote the empirical mean and standard deviation of the raw rewards across both online and offline samples in the current batch, as given by
\begin{align}
    \mu &= \frac{1}{G} \left(\sum_{x_0^i\in \mathcal{X}_{\text{online}}}s_{\text{raw}}^{(i)}+\sum_{x_0^j\in \mathcal{X}_{\text{offline}}}s_{\text{raw}}^{(j)}\right), \label{eq:mu} \\
    \sigma &= \sqrt{\frac{1}{N} \big[ \sum_{x_0^i\in \mathcal{X}_{\text{online}}} \left(s_{\text{raw}}^{(i)} - \mu\right)^2 + \sum_{x_0^j\in \mathcal{X}_{\text{offline}}}\left(s_{\text{raw}}^{(j)} - \mu\right)^2} \big]. \label{eq:sigma}
\end{align}
This normalization stabilizes training by reducing reward variance and preventing outlier domination. The resulting rewards $\{r^{(k)}\}$ are then used to update the policy via the objective in Eq. (\ref{eq:policy_opt}). This optimization encourages the denoising network to increase the likelihood of samples aligned with the VLM and human preference while suppressing opposite ones.

\begin{figure*}[h]
  \centering
  \includegraphics[width=\linewidth]{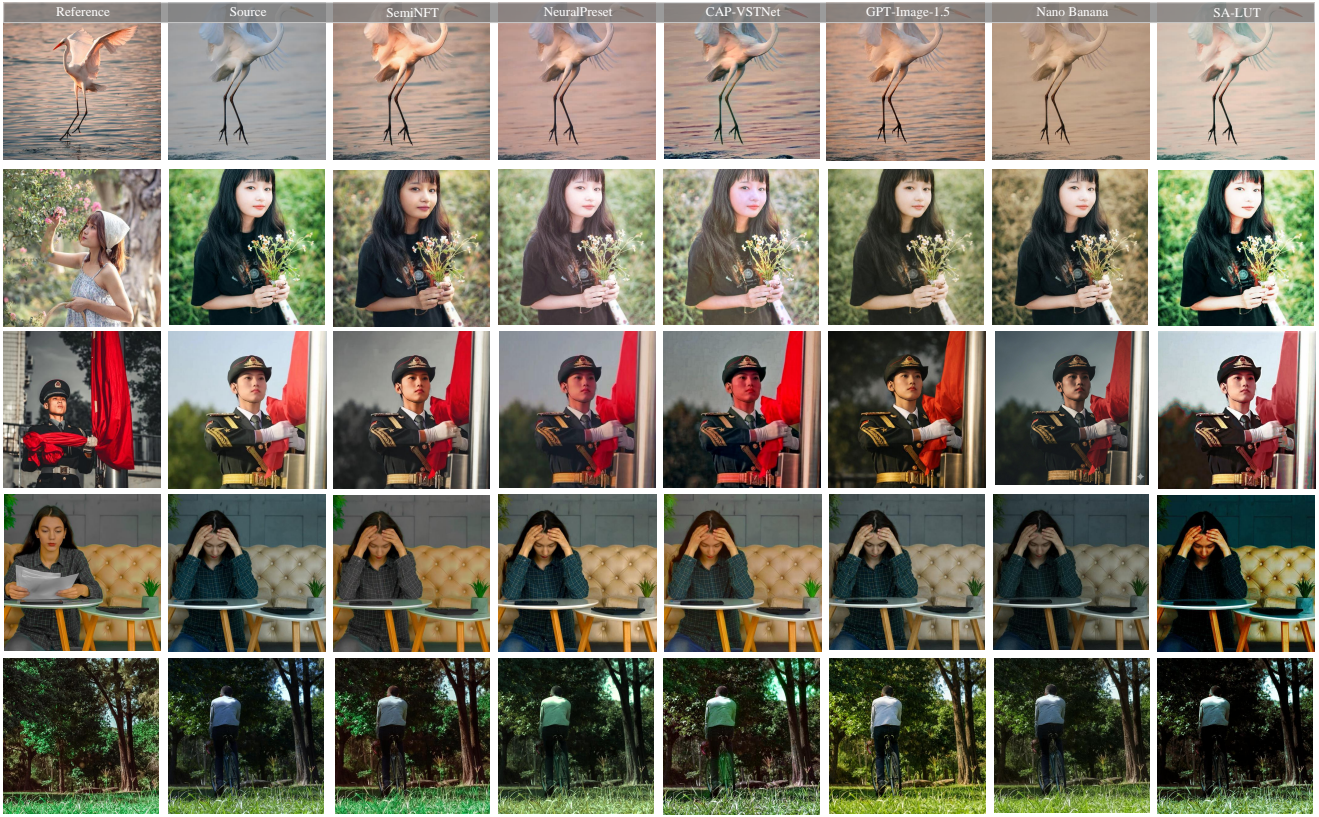}
  \caption{Visual comparison with baselines. The above three lines show the results on the realistic dataset, while the last two lines show results on the synthetic dataset.}
  \label{visual_main}
  \vspace{-0.5em}
\end{figure*}

\begin{figure*}[h]
  \centering
  \includegraphics[width=\linewidth]{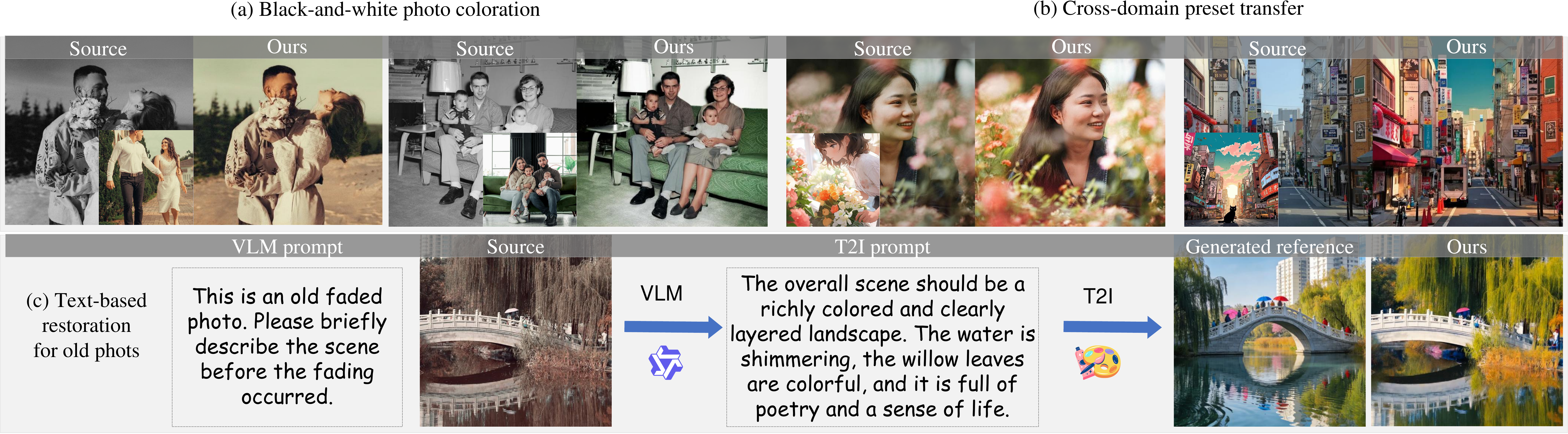}
  \caption{\textbf{Applications to extended scenes}, including black-and-white coloration, cross-domain preset transfer, and text-based restoration for old photos. (the down-left or down-right image is the reference image)}
  \label{visual_extended}
  \vspace{-0.5em}
\end{figure*}

\subsection{Dataset Construction}
\paragraph{Cold-Start dataset:} \label{cold-start dataset}
To support the cold-start training stage, we construct a paired dataset comprising quadruplets  $ (I_s, I_r, I_t, c) $ , where  $ I_s $  denotes the source image,  $I_r $  the reference image,  $ I_t $  represents the retouched target image, and  $c$ denotes  a unified textual description of the preset transfer task. We leverage an open-source video dataset \textit{Open-Sora-Plan-v1.1.0}, which contains $40,258$ videos covering portraits and nature landscapes. For each video, we randomly sample two temporally distinct frames and apply the same professionally designed preset for them, yielding $I_t$ and $I_r$. This ensures that  $ I_t $  and  $ I_r $  share the same stylistic preset while preserving distinct content. 
To better simulate related but non-identical real-world scenes, we randomly select $200$ video-derived pairs and use Qwen-Image-Edit \cite{qwenimage2025} as an instruction-guided editor to add or remove a salient object in the reference image (e.g., “add a red umbrella” or “remove the car”). This enhances robustness to content mismatch between source and reference. We carefully selected $50$ representative pairs as the evaluation set, named \textbf{synthetic dataset} while the rest were used as the training set.

\paragraph{RL dataset:} \label{RL dataset} 
For the RL stage, we construct an unpaired dataset of triplets  $ (I_s, I_r, c) $, where $ I_t $  is unavailable, which corresponds to the real-world conditions. Reference images  $ I_r $  are collected from multiple public photography communities\footnote{\url{https://pixabay.com/}, \url{https://www.pexels.com/}, \url{https://unsplash.com}}, selecting only high-resolution, well-composed photos with distinctive color grading. For each  $ I_r $, we perform cross-dataset image retrieval on the same sources using CLIP \cite{radford2021learning} embeddings to find a semantically related but visually distinct source image  $ I_s $. After filtering, the final RL Dataset contains \textbf{$1,500$} unpaired tuples. We carefully selected $150$ representative pairs as the evaluation set, named \textbf{realistic dataset} while the rest were used as the training set.

\begin{table*}[h]
\centering
\scriptsize  
\caption{Quantitative comparisons with baselines on \textbf{Realistic Dataset}. The best results are highlighted in \textbf{bold}.}
\renewcommand{\arraystretch}{1.05}
\setlength{\tabcolsep}{2.8pt}
\begin{tabular}{lcccc|c|c}
\toprule
\multirow{2}{*}{\textbf{Method}} & 
\multicolumn{4}{c|}{\textbf{VLM Score}} & 
\textbf{Content Similarity} &
\textbf{User Study} \\
& GPT-4o ↑ 
& Qwen3-VL ↑ & gemini-2.5-flash ↑ & Reward ↑ & Success Ratio ↑ 
& Ratio ↑ \\
\midrule
SA-LUT & 2.0467 & 2.1133 & 1.7633 & 0.3674 & 1.0 & - \\
NeurualPreset & 2.7667 & 2.404 & 2.2467 & 0.4271 & 1.0 & 0.1646\\
CAP-VSTNet & 2.6531 & 2.4853 & 2.2281 & 0.4304 & 1.0 & 0.0756 \\
PIF & 1.9214 & 2.1342 & 1.7321 & 0.3817 & 1.0 & - \\
\midrule
GPT-Image-1.5 & 2.6533 & 2.7245 & 2.7150 & 0.4420 & 1.0 & 0.1112 \\
Qwen-Image-Edit & 1.2473 & 1.3548 & 1.0752 & 0.2209 & 0.57 & - \\
Nano Banana & 2.2001 & 2.4132 & 2.1634 & 0.3975 & 0.95 & 0.1453 \\
\midrule
JarvisArt & 1.5321 & 1.7407 & 0.9655 & 0.3012 & 1.0 & - \\
\midrule
\textbf{Ours} & \textbf{3.6867} & \textbf{2.8639} & \textbf{3.5299} & \textbf{0.4931} & 1.0 & \textbf{0.5031} \\
\bottomrule
\end{tabular}
\label{tab:unpaired}
\end{table*}

\begin{table*}[h]
\centering
\scriptsize  
\caption{Quantitative comparisons with baselines on \textbf{Synthetic Dataset}. The best results are highlighted in \textbf{bold}.}
\renewcommand{\arraystretch}{1.05}
\setlength{\tabcolsep}{2.8pt}
\begin{tabular}{lcccc|cccc|c}
\toprule
\multirow{2}{*}{\textbf{Method}} & 
\multicolumn{4}{c|}{\textbf{Distortion-based metrics}} & 
\multicolumn{4}{c|}{\textbf{VLM Score}} & 
\textbf{User Study} \\
& PSNR ↑ & SSIM ↑ & DISTS ↓ & LPIPS ↓ & GPT-4o ↑ 
& Qwen3-VL ↑ & gemini ↑ & Reward ↑ 
& Ratio ↑ \\
\midrule
SA-LUT & 17.82 & 0.7460 & 0.1396 & 0.1688 & 2.9815 & 3.0556 & 2.5556 & 0.7687 & - \\
NeurualPreset & 22.16 & 0.8619 & 0.1092 & 0.1033 & 3.5556 & 3.6296 & 3.7407 & 0.8142 & 0.1849 \\
CAP-VSTNet & 23.41 & \textbf{0.8729} & 0.0971 & 0.0954 & 3.8519 & 3.6481 & 3.4444 & 0.8239 & 0.0548 \\
PIF & 17.94 & 0.8213 & 0.1460 & 0.1578 & 2.6296 & 3.0110 & 2.6667 & 0.7261 & - \\
\midrule
GPT-Image-1.5 & 18.70 & 0.6633 & 0.1439 & 0.1638 & 3.3333 & 3.6481 & 3.6661 & 0.8097 & 0.1199 \\
Qwen-Image-Edit & 13.62 & 0.4931 & 0.2212 & 0.3944 & 1.6130 & 1.7963 & 1.0232 & 0.3959 & - \\
Nano Banana & 16.13 & 0.5102 & 0.1506 & 0.2758 & 2.5741 & 3.7963 & 2.2222 & 0.8288 & 0.0822 \\
\midrule
JarvisArt & 14.48 & 0.6085 & 0.2261 & 0.3025 & 2.0192 & 2.4038 & 1.4231 & 0.6186 & - \\
\midrule
\textbf{Ours} & \textbf{25.68} & 0.8631 & \textbf{0.0743} & \textbf{0.0735} & \textbf{4.5185} & \textbf{3.9444} & \textbf{4.3333} & \textbf{0.8683} & \textbf{0.5581} \\
\bottomrule
\end{tabular}
\label{tab:paired}
\end{table*}


\section{Experiments}
\subsection{Experimental Setup}
\paragraph{Evaluation Metrics:}
We evaluate our method on both the \textbf{realistic dataset} and the \textbf{synthesized dataset}.
On the realistic dataset, we assess performance through three complementary lenses:
(i) \textbf{VLM-based evaluation}: We use three VLMs, GPT-4o, Qwen3-VL-32B, and Gemini-2.5-Flash, to score the global and local color similarity to the reference image’s color. Each criterion is rated on a 5-point Likert scale accompanied by qualitative justifications. We also include the logit-based reward model.
(ii) \textbf{Success ratio}: Several baseline methods based on generative models occasionally fail to follow instructions, either by directly outputting the reference image, which yields artificially high VLM scores due to identical-image comparison, or by generating entirely new content. To quantify such catastrophic content distortion, we define a binary success metric based on the DINO feature similarity \cite{caron2021emerging} between the source and output images. An output is classified as successful if the similarity exceeds a predefined threshold. The success ratio is reported as the percentage of successful cases.
(iii) \textbf{User study}: 
We conduct a human evaluation that focuses on the color similarity to the reference image and subjective aesthetic. Participants are asked to select the best result among competing methods. Average scores are reported, with additional details provided in Appendix~\ref{appendix: user}.
On the synthesized dataset, we additionally compute standard distortion metrics against the  target image: PSNR, SSIM \cite{wang2004image}, LPIPS \cite{zhang2018unreasonable}, and DISTS \cite{ding2020image}, including pixel-level evaluation and feature-level evaluation.

\paragraph{Implementation Details:}
We employ FLUX.1-dev as the pre-trained DiT backbone. 
In the cold-start training stage, we train for approximately $10,000$ iterations, which takes about $24$ hours using $2$ NVIDIA H20-96G GPUs.
In the RL stage, we set the group size to $G=9$. To accelerate policy rollouts, we use only $T=6$  denoising steps during training, while evaluations are performed with $T\!=\!28$ steps for higher fidelity. The entire RL stage converges in approximately $150$ GPU-hours.

\subsection{Comparison with existing methods}
We compare SemiNFT with four SOTA photorealistic preset transfer methods,
including SA-LUT\cite{Gong2025salut}, Neural Preset\cite{ke2023neural}, PIF\cite{zhu2025personalized}, and CAP-VSTNet \cite{wen2023cap}, one agent-based method JarvisArt \cite{jarvisart2025}, and three unified image editing methods, including GPT-Image-1.5, Nano Banana, and Qwen-Image-Edit. Detailed descriptions of the baseline implementations are provided in Appendix \ref{appendix: baseline}.

\paragraph{Realistic Dataset:} The quantitative results on the realistic dataset are shown in Tab. \ref{tab:unpaired}, from which we observe that our method achieves SOTA performance in terms of most metrics. The visual results are provided in Fig. \ref{visual_main}, where we compare representative methods with strong quantitative results. Specifically, SA-LUT tends to introduce severe aliasing artifacts and over-sharpening, while failing to achieve accurate color alignment at both global and local levels. Neural Preset and CAP-VSTNet typically sample colors in the vicinity of the corresponding area in the source image; hence, the generated results can be unsatisfactory when there is a large discrepancy between the reference and the source. Furthermore, due to their lack of generative priors, these methods underperform in complex scenarios like portrait retouching, which demands deep awareness of semantic context, often leading to color bleeding (e.g., assigning sky blue to facial regions) and unintended structural distortions.
Among generative approaches, we prompt GPT-Image-1.5 using a carefully designed multi-turn conversation template to guide it toward the preset transfer task, enabling it to produce visually appealing outputs. For Gemini-2.5-Image, generation is performed via a single-turn instruction. Full prompt templates are provided in Appendix \ref{appendix: baseline}. However, neither model supports fine-grained semantic alignment; both produce coarse colorization results, with poor local details.

\paragraph{Synthetic Dataset:} The quantitative results on synthetic dataset are shown in Tab. \ref{tab:paired}, and the qualitative results are shown in Fig. \ref{visual_main}. It can be found that our method achieves the best performance among nearly all metrics, demonstrating the performance of preset transfer and content preservation.

\begin{table}[h]
\centering
\scriptsize  
\caption{Ablation Studies. The best results are highlighted in \textbf{bold}.}
\renewcommand{\arraystretch}{1.05}
\setlength{\tabcolsep}{2.8pt}
\begin{tabular}{lccc|c}
\toprule
\multirow{2}{*}{\textbf{Method}} & 
\multicolumn{3}{c|}{\textbf{VLM Score}} &
\multicolumn{1}{c}{\textbf{Content}} \\ 
& GPT-4o ↑ 
& Qwen3-VL ↑ & Reward ↑ & $\text{DINO}_{\text{local}}$ ↑ \\
\midrule
SemiNFT & 3.6867 & \textbf{2.8639} & 0.4931 & \textbf{0.9948} \\
\midrule
w/o cold-start stage & 0.3111 & 0.4090 & 0.1245 & 0.0 \\
w/o RL stage & 2.9185 & 2.4030 & 0.4196 & 0.9843\\
w/o realistic dataset & 3.5102 & 2.7998 & 0.4869 & 0.9633\\
w/o offline samples & \textbf{3.6999} & 2.8377 & \textbf{0.4954} & 0.9738\\
\bottomrule
\end{tabular}
\label{tab:ablation}
\end{table}
 \subsection{Extended Scenarios}
Our method demonstrates strong zero-shot generalization to a variety of challenging tasks, despite the absence of any task-specific data incorporated in the training stage.

\paragraph{Black-and-white Photo Colorization:}
As illustrated in Fig. \ref{visual_extended}, SemiNFT achieves impressive colorization results. Although a significant chromatic discrepancy exists between the black-and-white source image and color reference image, our model facilitates precise color mapping by aligning semantic regions, ensuring skin-to-skin and background-to-background consistency, demonstrating spatially coherent and semantically aware color transfer.

\paragraph{Cross-Domain Preset Transfer:} 
The robustness of SemiNFT is further evidenced by its capacity for cross-domain preset transfer, such as translating aesthetics between anime images and realistic photographs. This scenario is particularly challenging due to the significant domain gap in texture, shading, and detail density. 
Unlike conventional style transfer that often alters the underlying content, SemiNFT strictly preserves the source content while changing only the color and tonal characteristics.
This capability provides immense value for creative industries, enabling artists to bridge disparate artistic mediums and apply stylized aesthetics to real-world content with unprecedented structural fidelity.

\paragraph{Text-guided Restoration for Old Photos:} 
Restoring vintage photos is often hindered by the difficulty of discerning the original colors and finding photorealistic references that match faded scenarios.
To address this, SemiNFT can be seamlessly integrated with existing VLMs and Text-to-Image (T2I) generative models. Specifically, by utilizing VLM to infer the possible color of the source image, a T2I model can synthesize a high-fidelity reference image for restoration. This automated pipeline transforms the restoration task from a search for rare references into a generative process, showcasing the intelligent adaptability of our model.

\subsection{Ablation Study}


Table \ref{tab:ablation} presents ablation study results that assesses our training strategies, dataset design, and reward mechanisms. While the primary metrics assess transfer quality, we include a content preservation metric to monitor for reward hacking. Specifically, we define a binary $\text{DINO}_{\text{local}}$ metric based on DINO \cite{caron2021emerging} similarity between source and output; samples with similarity below a threshold are counted as `local change'.


The results demonstrate that the cold-start stage establishes an essential structural foundation for processing multi-input source and reference image pairs, a capability that is not natively supported by the base FLUX.1-[dev] model. The RL stage plays a critical role in local refinement, as its removal leads to degradation in semantic consistency and content fidelity. In addition, the hybrid data pipeline is crucial for real-world generalization, as models trained exclusively on synthetic data fail to transfer effectively to realistic images. Finally, the hybrid online–offline reward mechanism substantially mitigates reward hacking. Without this mechanism, the policy tends to over-optimize transfer-related rewards at the expense of alignment with the source structure. More qualitative comparisons that support these observations are provided in Appendix~\ref{appendix: ablation}.

\section{Conclusion}
In this paper, we presented SemiNFT, a novel DiT-based framework for professional color retouching. By implementing a progressive curriculum-style training strategy, the model transitioned from supervised pretraining on paired dataset to reinforcement learning on unpaired datasets, effectively cultivating both basic structural preservation, mapping skills and nuanced aesthetic perception. Furthermore, to mitigate reward hacking and the risk of catastrophic forgetting during the RL stage, we introduced a hybrid online-offline reward mechanism, which ensures structural integrity while pursuing aesthetic optimization. Our experimental results demonstrated that SemiNFT consistently outperformed state-of-the-art methods across standard benchmarks. Furthermore, the model's superior performance in zero-shot scenarios—including black-and-white colorization and cross-domain preset transfer, demonstrating its potential wide applications.

\bibliographystyle{ieeenat_fullname}
\bibliography{main}

\newpage
\appendix
\onecolumn

\section{User Study} \label{appendix: user}
\begin{figure*}[h]
  \centering
  \includegraphics[width=\linewidth]{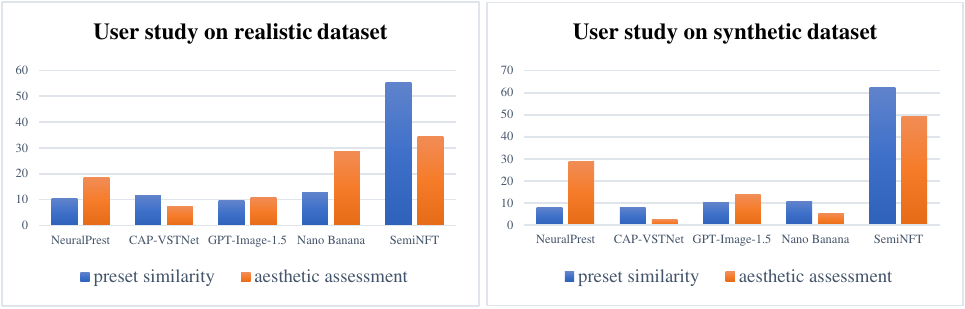}
  \caption{\textbf{User Study: }Selection ratios of preset similarity and human aesthetics on realistic and synthetic dataset.}
  \label{user_study}
  \vspace{-0.5em}
\end{figure*}
We conduct a user study to judge whether the generated images of SemiNFT align well with human perception. A total of $72$ participants took part in the study. We compared SemiNFT with four strong baselines, including Neural Preset, CAP-VSTNet, GPT-Image-1.5, and Nano Banana. Our assessments include the preset similarity and the subjective aesthetics. For each question, participants were provided with a reference image, the source image, and multiple results. They were asked to select the best result of preset similarity to the reference and the one that fits their preference most, respectively. During the analysis, each selection made for a particular model was counted as one point, and the percentage score for each model was calculated based on the total number of selections, yielding selection ratios. The selection ratios are shown in Fig. \ref{user_study}, indicating that our SimiNFT results received higher user preference in terms of both preset similarity and subjective aesthetics.

\section{Ablation Study Visualization} \label{appendix: ablation}
\begin{figure*}[h]
  \centering
  \includegraphics[width=\linewidth]{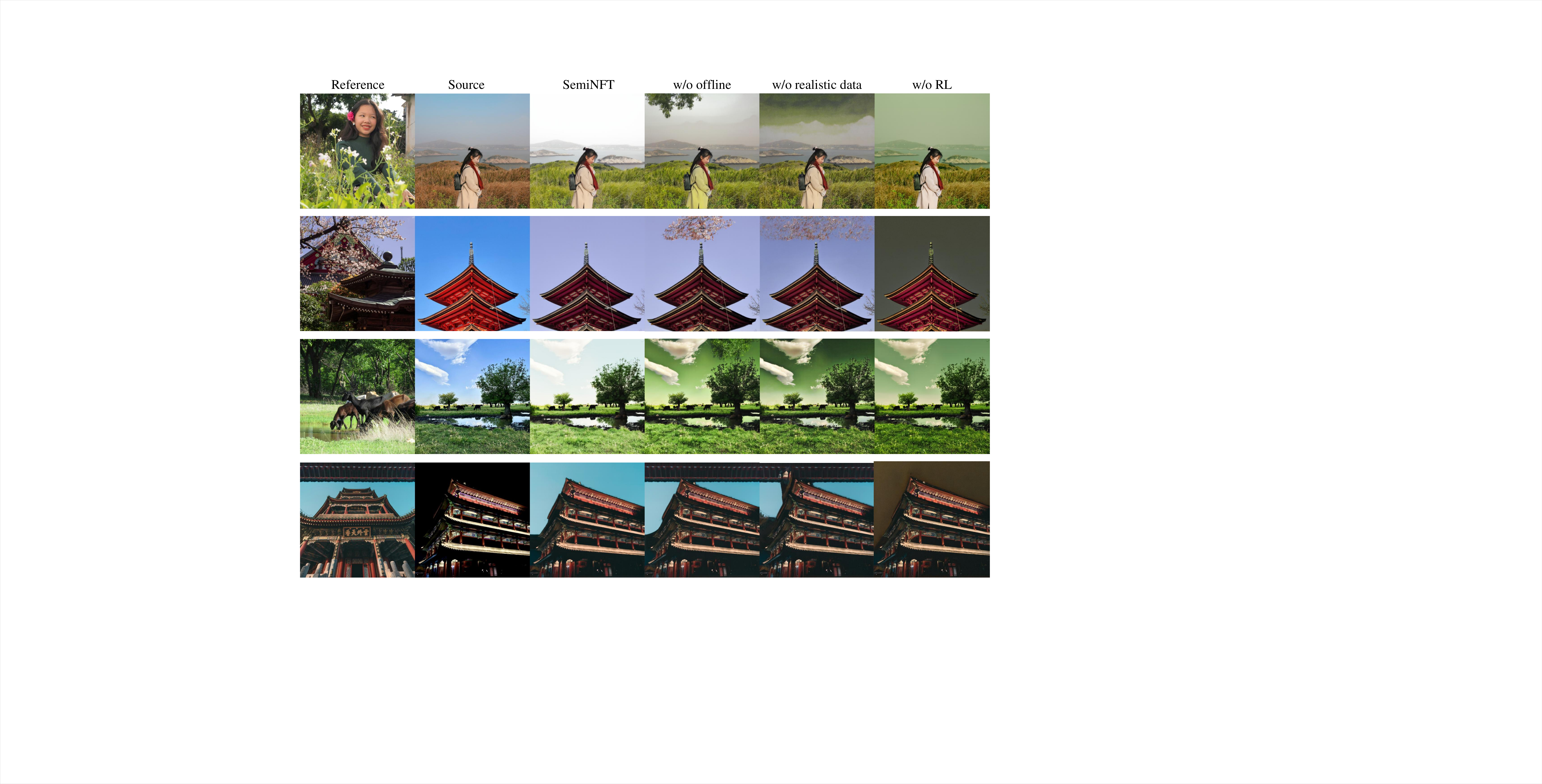}
  \caption{\textbf{Visualization of ablation studies.} Here we include SemiNFT, `w/o offline samples', `w/o realistic dataset', and `w/o RL'. For clarity, we omit `w/o cold-start stage' as it's the worst.}
  \label{visual_ablation}
  \vspace{-0.5em}
\end{figure*}

We visualize the results of our ablation studies in this section. For clarity, we omit ``SemiNFT w/o cold-start stage”, as it suffers from catastrophic failures, producing completely unrelated outputs.
As shown in Fig. \ref{visual_ablation}, ``SemiNFT w/o RL stage” fails to accurately transfer the reference preset, where the sky is rendered in a color unrelated to that of the reference.
The variant ``SemiNFT w/o realistic dataset” exhibits poor content preservation, often generating  unexpected elements. This is because the synthetic training data contains highly aligned source–reference pairs, limiting the model’s ability to generalize to real-world scenarios with greater domain gaps.
Similarly, ``SemiNFT w/o offline samples” tends to copy visual elements directly from the reference image—not only its color palette but also its content—thereby disrupting the source structure. Besides, the generator overfits to the color distribution of the reference at the expense of photorealism, generating unnatural results.
In contrast, our full SemiNFT model achieves both accurate preset transfer and faithful content preservation, yielding outputs that align well with human aesthetic preferences, showing the advantages of the curriculum-style training pipeline, dataset construction, and the hybrid reward mechanism.

\section{Baseline Setup}  \label{appendix: baseline}
\subsubsection{GPT-Image-1.5}
\begin{figure*}[h] 
\centering
\begin{promptbox}[Template of multi-turn conversation for GPT-Image-1.5 generation] 
\textbf{\color{blue} User:} I invite you to participate in an experiment on ’Image Preset Filter Transfer’. Be patient and focus. I will pay you tips according to your performance. Are you ready now?

\textbf{\color{red} Assistant:} Locked in, patient, and focused. Let’s do the experiment.

\textbf{\color{blue} User:} Step 1: I will provide a reference image. Please first view the image.

\textbf{\color{red} Assistant:} Got it. Please go ahead and share the reference image—I’m ready to view it carefully.

\textbf{\color{blue} User:}  (\textit{upload the reference image})

\textbf{\color{red} Assistant:}  I’ve viewed the reference image carefully. I see a person positioned low among dense green foliage...

\textbf{\color{blue} User:} Step 2: I will provide an input image. Please transfer the filter (photographic preset filter) of the reference image to the input image. Give me the result image.

\textbf{\color{red} Assistant:} Understood.
Please share the input image, and I’ll transfer the photographic preset/filter from the reference image onto it carefully and consistently.

\textbf{\color{blue} User:}  (\textit{upload the source image})

\textbf{\color{red} Assistant:}  (\textit{call GPT-Image-1.5 to generate an image based on the above conversation})
\end{promptbox}

\begin{promptbox}[Text Prompts for single-turn Gemini-2.5-flash-image generation] 
You will receive two images as input. The first image is the source image, and the second image is the reference image. Please transfer the color from the reference image onto the source image to make the overall color atmosphere of the two images consistent, and the colors of the corresponding elements in the local areas should also remain consistent. Note: Only change the colors of the source image, and do not alter its content structure.
\end{promptbox}

\includegraphics[width=\textwidth]{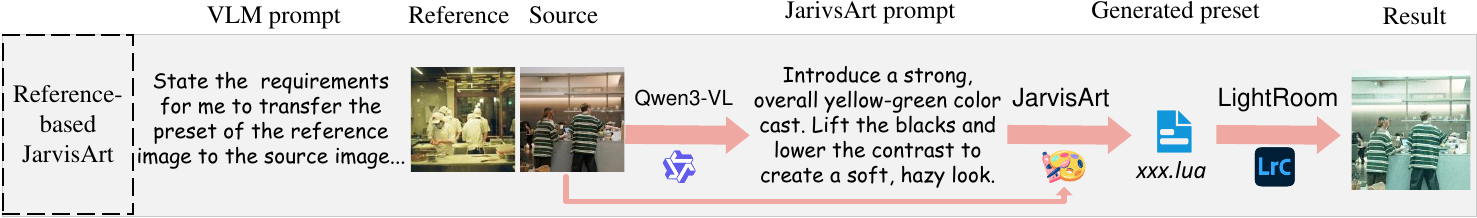} 

\caption{Template for GPT-Image-1.5 generation and Gemini-2.5-flash-image generation, and the pipeline for reference-based JarvisArt.}
\label{templet_gpt_gemini}
\end{figure*}

We prompt GPT-Image-1.5 using a carefully designed multi-turn conversation template, as shown in Fig. \ref{templet_gpt_gemini}, to guide it toward the preset transfer task, enabling it to produce visually appealing outputs. 

\subsubsection{Nano Banana}
For Gemini-2.5-Image, generation is performed via a single-turn instruction. Full prompt templates are provided in Fig. \ref{templet_gpt_gemini}.

\subsubsection{Javisart}
JarvisArt is the SOTA agent-based retouching model. To enable a broad comparison, we include it as a baseline. Since JarvisArt operates on text instructions, it requires a textual description of the retouching requirements. To generate this prompt, we employ the Qwen3-VL-32B to analyze the source and reference images and produce an appropriate textual instruction. The complete pipeline is illustrated in Fig. \ref{templet_gpt_gemini}.


\section{Temples Design for Reward Model and Evaluation} \label{template}
In this section, we present the templates used for the reward model, as illustrated in Fig. \ref{templet_reward}. During training, the model is constrained to output a single numerical score to enable efficient scoring. For evaluation, however, we employ the same templates to generate structured JSON-formatted outputs that include both a score and a corresponding analysis, facilitating explainability assessment.

\begin{figure*}[h] 
\centering
\begin{promptbox}[Template for Reward Model] \label{templet}

\textbf{System Prompt:} You are an extremely rigorous ``Holistic and Local Color Harmony Specialist." Your core responsibility is to evaluate the \textbf{visual color similarity} between two input images, considering both the \textbf{overall atmosphere} and the \textbf{color consistency of key local elements}. You strive for professional, quantified, and balanced color matching.

\textbf{User Prompt:} 
\\ \\
\textbf{Task} \\
Your sole task is to determine the matching degree between the two input images across three dimensions: \textbf{overall color tone tendency, color saturation, and brightness distribution}. When deriving the final score, you must internally weigh the \textbf{similarity of the overall atmosphere} against the \textbf{color discrepancies of key local elements}. Ignore subjective aesthetics and compositional factors; base your score (0-5) strictly on professional colorimetry standards.
\\ \\
\textbf{Scoring Criteria (Mandatory)}
\begin{itemize}
    \item 0 (Absolute Difference): At least two dimensions among main tone, saturation, or brightness are completely opposite or entirely unrelated.
    \item 1 (Extremely Low Match): Fundamental differences exist in the overall color style or the color tendency of most key local elements.
    \item 2-3 (Moderate Match): Overall color tone tendencies are similar, but significant differences exist in \textbf{saturation or brightness}, or the color treatment of a few key local elements is clearly inconsistent.
    \item 4 (High Match): The overall color style is highly unified, and the color treatment of most key local elements (e.g., sky, skin tones, etc.) is also highly consistent.
    \item 5 (Perfect Match): Almost no perceptible difference across \textbf{all key color dimensions} and \textbf{all visible local elements}, reaching the color consistency standards of professional commercial photography series.
\end{itemize}
\textbf{Important}\\
Please note: \\
1. \textbf{Overall Atmosphere Priority}: The final score is based on the overall color atmosphere, but discrepancies in local elements will act as deductions affecting the final rating. \\
2. \textbf{Focus Only on Color}: Completely ignore objects, composition, subject matter, and image clarity.\\

\textbf{Response Format:}

Output ONLY a single number from 0 to 5. (\color{red} reward model) \color{black}

Output form:
{
"score" : [...],
"reasoning" : "..."
}. (\color{blue}evaluation) \color{black}
\end{promptbox}

\caption{Template for the reward model is forced to output a single number for fast training, while for evaluation, it is to output a score with corresponding reasons.}
\label{templet_reward}
\end{figure*}


\section{More visualizations on Realistic Dataset}
\begin{figure*}[h]
  \centering
  \includegraphics[width=\linewidth]{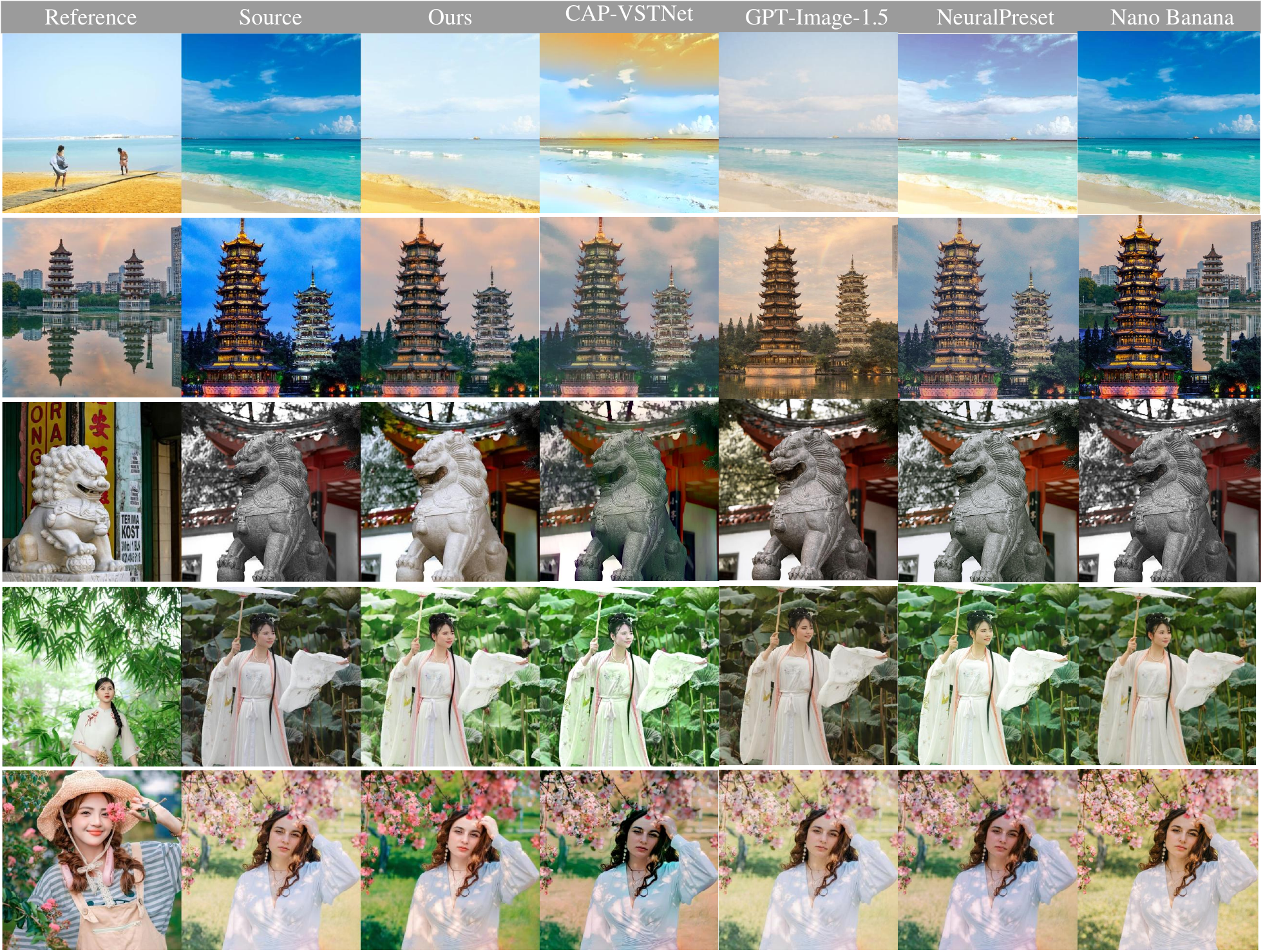}
  \caption{More visualization results on the proposed realistic dataset.}
  \label{more_visual}
  \vspace{-0.5em}
\end{figure*}

\begin{figure*}[t]
  \centering
  \includegraphics[width=\linewidth]{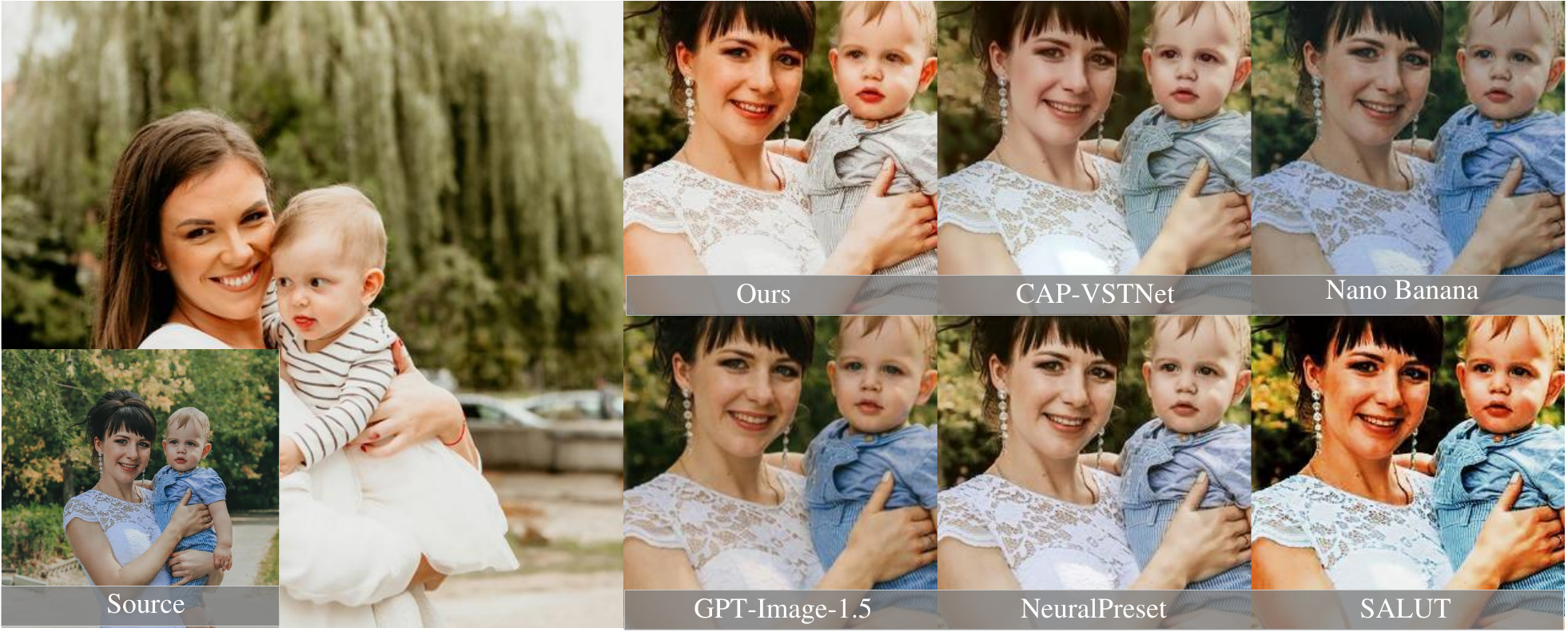}
  \caption{Local detail comparisons.}
  \label{detail_visual}
  \vspace{-0.5em}
\end{figure*}

\begin{figure*}[t!]
  \centering
  \includegraphics[width=\linewidth]{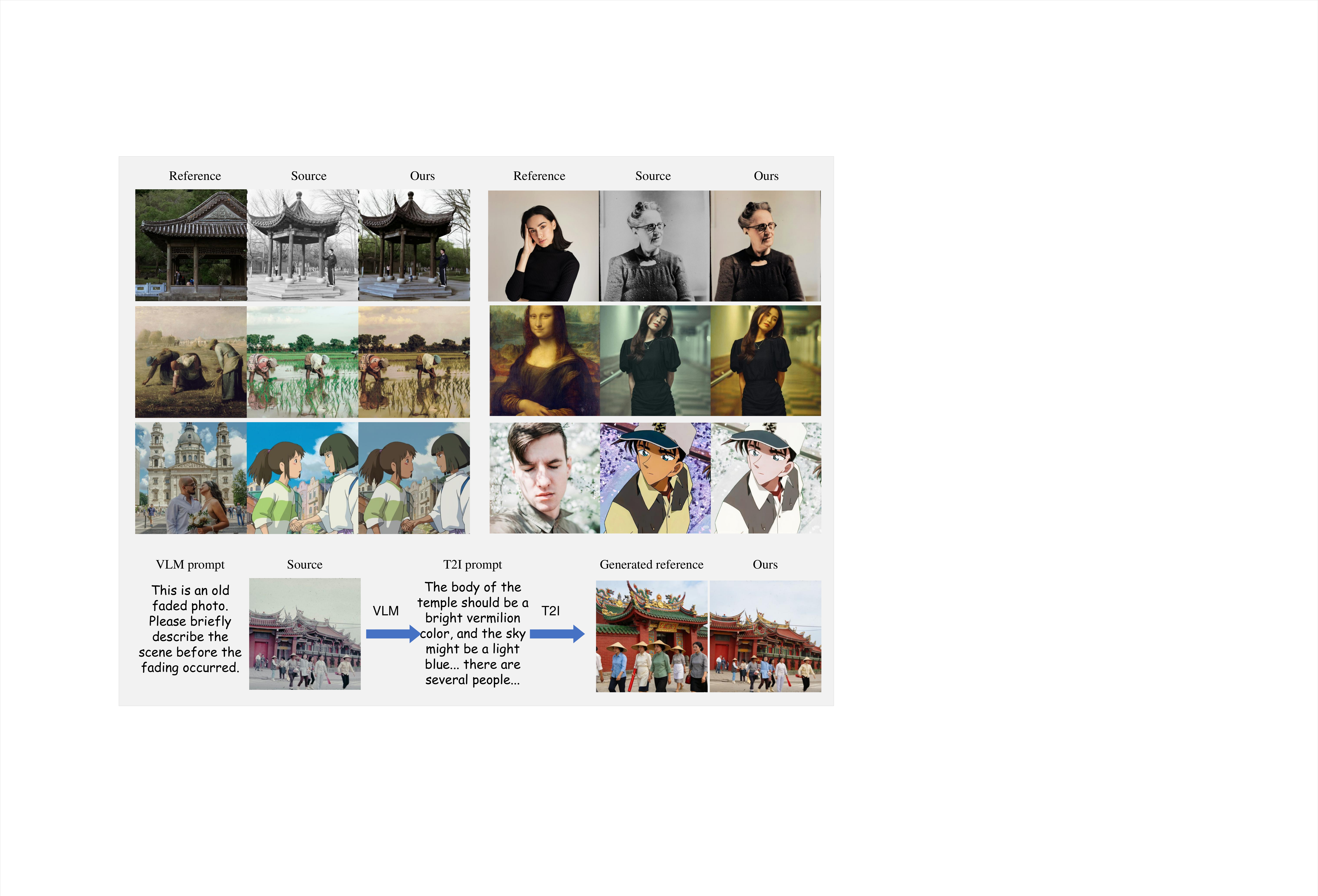}
  \caption{\textbf{More visualization results on applications to extended scenes, including black-and-white coloration, cross-domain preset transfer, and text-based restoration for old photos.}}
  \label{visual_extended_appendix}
  \vspace{-0.5em}
\end{figure*}

In this section, we present additional visualization results on the  proposed realistic dataset. As shown in Fig. \ref{more_visual}, our model is compared against four baseline methods. Furthermore, a detailed local comparison is provided in Fig. \ref{detail_visual}, clearly demonstrating that our approach achieves precise local skin texture transfer, which has gone beyond global atmosphere. Additionally, Fig. \ref{visual_extended_appendix} showcases more results on the extended scenarios, including black-and-white colorization, cross-domain preset transfer, and text-guided restoration of old photographs, showing that our model transcends simple statistical matching.

\section{Results on PST Dataset}
\begin{figure*}[t!]
  \centering
  \includegraphics[width=\linewidth]{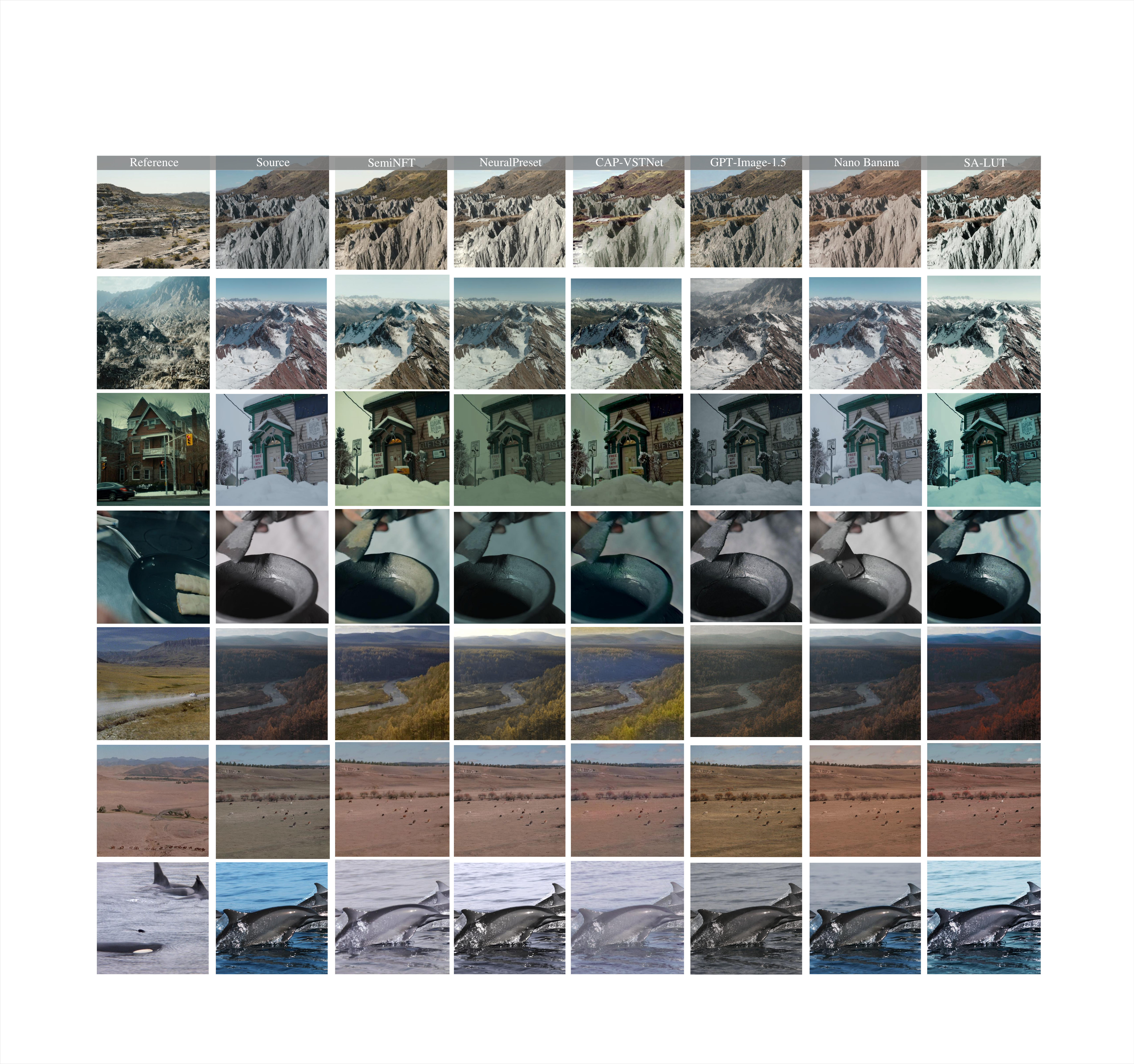}
  \caption{Visual comparisons on the PST50-unpaired dataset.}
  \label{visual_pst_unpaired}
  \vspace{-0.5em}
\end{figure*}

\begin{figure*}[h]
  \centering
  \includegraphics[width=\linewidth]{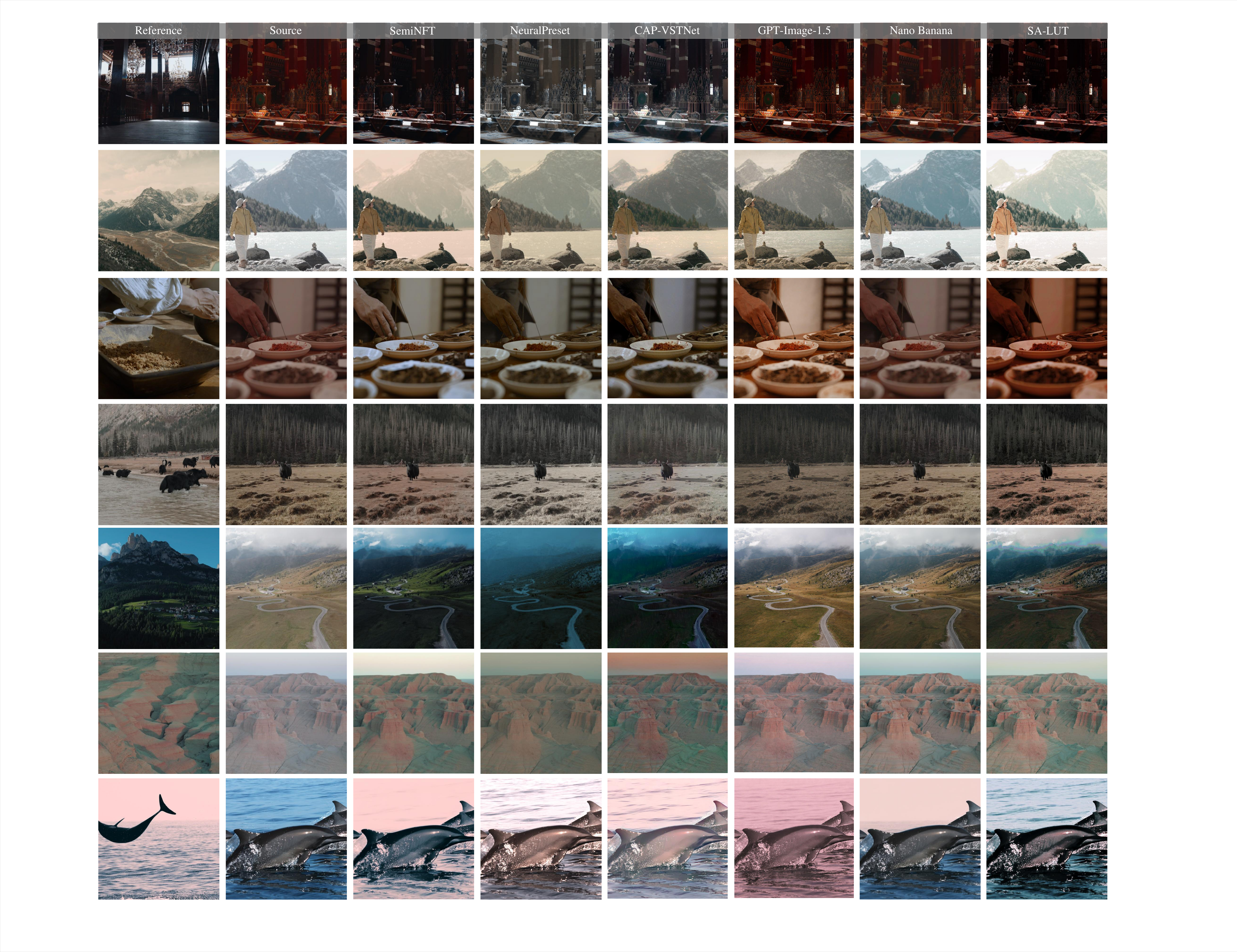}
  \caption{Visual comparisons on the PST50-paired dataset.}
  \label{visual_pst_paired}
  \vspace{-0.5em}
\end{figure*}

To evaluate the generalization capability of SemiFNT, we present results on another dataset-PST50 dataset\cite{Gong2025salut}. We compare against strong baselines, including NeuralPreset, CAP-VSTNet, GPT-Image-1.5, Nano Banana, and SA-LUT. For brevity, we report only the main metrics, as they exhibit high correlation. Quantitative results are provided in Tab. \ref{tab:pst_unpaired} and Tab. \ref{tab:pst_paired}, corresponding to the PST50-unpaired and PST50-paired subsets, respectively. Following our previous evaluation protocol, for the unpaired setting, we assess performance using VLM-based metrics and content similarity measured by success ratio; for the paired setting, we additionally incorporate distortion-based metrics. The results demonstrate that SemiFNT performs consistently well across all metrics on both datasets, highlighting its strong generalization ability. Visualization results for the PST50-unpaired and PST50-paired settings are shown in Fig. \ref{visual_pst_unpaired} and Fig. \ref{visual_pst_paired}, respectively.

\begin{table*}[t!]
\centering
\scriptsize  
\caption{Quantitative comparisons with baselines on \textbf{PST50-unpaired dataset}. The best and second best results are highlighted in \textbf{bold}.}
\renewcommand{\arraystretch}{1.05}
\setlength{\tabcolsep}{2.8pt}
\begin{tabular}{lccc|c}
\toprule
\multirow{2}{*}{\textbf{Method}} & 
\multicolumn{3}{c|}{\textbf{VLM Score}} & 
\textbf{Content Similarity}  \\
& GPT-4o ↑ 
& Qwen3-VL ↑ & Reward ↑ & Success Ratio ↑ \\
\midrule
SA-LUT & 2.1800 & 2.1600 & 0.3170 & 1.0 \\
NeurualPreset & 2.7600 & 2.6400 & 0.3845 & 1.0 \\
CAP-VSTNet & 3.1000 & 2.6000 & 0.3827 & 1.0 \\
\midrule
GPT-Image-1.5 & 2.9200 & 2.6400 & 0.3827 & 1.0 \\
Nano Banana & 2.1000 &2.4000 & 0.3149 & 0.96 \\
\midrule
\textbf{Ours} & \textbf{3.5800} & \textbf{2.8400} & \textbf{0.4249} & 1.0 \\
\bottomrule
\end{tabular}
\label{tab:pst_unpaired}
\end{table*}

\begin{table*}[h]
\centering
\scriptsize  
\caption{Quantitative comparisons with baselines on \textbf{PST50-paired dataset}. The best and second best results are highlighted in \textbf{bold}.}
\renewcommand{\arraystretch}{1.05}
\setlength{\tabcolsep}{2.8pt}
\begin{tabular}{lcccc|ccc}
\toprule
\multirow{2}{*}{\textbf{Method}} & 
\multicolumn{4}{c|}{\textbf{Metrics with GT}} & 
\multicolumn{3}{c|}{\textbf{VLM Score}} \\
& PSNR ↑ & SSIM ↑ & DISTS ↓ & LPIPS ↓ & GPT-4o ↑ 
& Qwen3-VL ↑ & Reward ↑ \\
\midrule
SA-LUT & 19.13 & 0.7661 & 0.1530 & 0.1386 & 3.1800 & 2.7200 & 0.5503 \\
NeurualPreset & 19.25 & 0.7968 & 0.1248 & 0.1205 & 3.3000 & 2.9000 & 0.6159 \\
CAP-VSTNet & 19.94 & 0.8086 & 0.1487 & 0.1174 & 3.8800 & 3.1000 & 0.6576 \\
\midrule
GPT-Image-1.5 & 19.77 & 0.7061 & 0.1411 & 0.1502 & 3.4000 & 2.9400 & 0.6102 \\
Nano Banana & 19.85 & 0.7182 & 0.1386 & 0.1513 & 3.2000 & 2.9200 & 0.5906 \\
\midrule
\textbf{Ours} & \textbf{20.64} & \textbf{0.8277} & \textbf{0.1166} & \textbf{0.1133} & \textbf{4.1200} & \textbf{3.3600} & \textbf{0.7328} \\
\bottomrule
\end{tabular}
\label{tab:pst_paired}
\end{table*}

\end{document}

%% file: preamble.tex
\usepackage{lineno}
\usepackage{multirow}
\usepackage{pifont}
\usepackage{graphicx}
\usepackage[dvipsnames]{xcolor}
\usepackage{amsmath}
\usepackage{amssymb}
\usepackage{booktabs}
\usepackage{overpic}
\usepackage{float}
\usepackage[linesnumbered, boxed, ruled]{algorithm2e}
\usepackage{array}
\usepackage{soul}
\usepackage{makecell}
\usepackage[table]{xcolor}
\usepackage{tcolorbox}
\usepackage{balance}
\usepackage[normalem]{ulem}
\useunder{\uline}{\ul}{}
\usepackage{leftindex}
\usepackage{arydshln}
\usepackage{enumitem}
\usepackage{siunitx}

\usepackage{bibunits}
\defaultbibliographystyle{ieeenat_fullname}
\defaultbibliography{main}